\pdfoutput=1

\documentclass[11pt]{article}

\usepackage{acl}

\usepackage{times}
\usepackage{latexsym}
\usepackage{booktabs}
\usepackage{amsmath}
\usepackage{algorithm}
\usepackage{algpseudocode}
\usepackage{graphicx}
\usepackage{subcaption}
\usepackage{multirow}
\usepackage{tabularx}
\usepackage{enumitem}
\usepackage{svg}
\usepackage{tabularray}
\usepackage{longtable}
\usepackage{amsthm}
\usepackage{bm}
\usepackage{stackengine}

\usepackage{cleveref}

\setitemize{labelindent=0em,labelsep=0.2cm,leftmargin=*}
\DeclareFontFamily{U}{mathx}{\hyphenchar\font45}
\DeclareFontShape{U}{mathx}{m}{n}{%
<-6> mathx5
<6-7> mathx6
<7-8> mathx7
<8-9> mathx8
<9-10> mathx9
<10-12> mathx10
<12-> mathx12
}{}

\DeclareSymbolFont{mathx}{U}{mathx}{m}{n}
\DeclareFontSubstitution{U}{mathx}{m}{n}

\DeclareMathSymbol{\bigovoid}{\mathop}{mathx}{"EC}

\usepackage{xcolor}
\DeclareMathOperator*{\softmax}{softmax}

\DeclareMathOperator*{\argmax}{argmax}

\usepackage[T1]{fontenc}

\usepackage[utf8]{inputenc}

\usepackage{microtype}

\usepackage{inconsolata}

\title{Towards Low-Resource Alignment to Diverse Perspectives with Sparse Feedback}

\author{Chu Fei Luo\textsuperscript{\hspace{0.5mm}1,2}, \textbf{Samuel Dahan\textsuperscript{\hspace{0.5mm}2,3}, and Xiaodan Zhu\textsuperscript{\hspace{0.5mm}1,2}}\\
\textsuperscript{1}Department of Electrical and Computer Engineering \& Ingenuity Labs Research Institute \\ Queen's University\\
\textsuperscript{2}Conflict Analytics Lab, Queen's University\\
\textsuperscript{3}Cornell Law School\\
\small{\{\texttt{chufei.luo,samuel.dahan,xiaodan.zhu}\}\text{\texttt{@queensu.ca}}}}

\begin{document}
\maketitle
\begin{abstract}
As language models have a greater impact on society, it is important to ensure they are aligned to a diverse range of perspectives and are able to reflect nuance in human values. However, the most popular training paradigms for modern language models often assume there is one optimal answer for every query, leading to generic responses and poor alignment. In this work, we aim to enhance pluralistic alignment of language models in a low-resource setting with two methods: pluralistic decoding and model steering. We empirically demonstrate that model steering offers consistent improvement over zero-shot and few-shot baselines with only 50 annotated samples. Our proposed methods decrease false positives in several high-stakes tasks such as hate speech detection and misinformation detection, and improves the distributional alignment to human values in GlobalOpinionQA. We hope our work highlights the importance of diversity and how language models can be adapted to consider nuanced perspectives.
\footnote{Our code is available at \url{https://github.com/chufeiluo/SAE-PD}}
\end{abstract}
\section{Introduction}

\vspace{-1mm}
Recent advancements in natural language processing, driven by Reinforcement Learning from Human Feedback (RLHF), have garnered a significant amount of interest \cite{ouyang2022training}. Experts have begun adopting Large Language Models (LLMs) in applications with increasing social impact. With this rapid adoption, there are many emerging challenges with \emph{AI alignment}, ensuring that automated systems are developed in the best interest of humans. The very definition of ``best interest'' is nuanced and open to exploration. In both reinforcement learning and fine-tuning settings, traditional machine learning conventions assume that there is one optimal answer \cite{kirk2023personalisation, poddar2024personalizing}. In cases of extreme diverging preferences, this causes a reward model to prefer generic responses that do not fully satisfy anyone \cite{poddar2024personalizing}.
Guiding language models to reflect multiple perspectives, also known as pluralistic alignment \cite{sorensen2024roadmap}, is essential for high-stakes tasks in law, medicine, and finance that are often dependent on diverse preferences.

AI alignment is an open-ended research question with significant room for exploration, and inter-disciplinary collaboration is essential for ensuring proper representation of a task's diversity \cite{wang2023aligning}.
One advantage to LLMs is the ability to improve the system with in-context instructions, which allows for more semantically rich feedback from domain experts and facilitates stronger collaboration.
It is also important to consider the cost of alignment --- training strategies are resource-intensive in both compute and annotations, which is often expensive for domain-specific tasks. In practice, a domain expert also cannot provide feedback for every sample at inference time. We wish to explore a \textit{sparse setting} with minimal data, where a domain expert can quickly specify their perspective in a few samples. 

\begin{figure*}
    \centering
    \includegraphics[trim=0.5cm 0.5cm 0.5cm 0.5cm, clip, width=0.75\linewidth ]{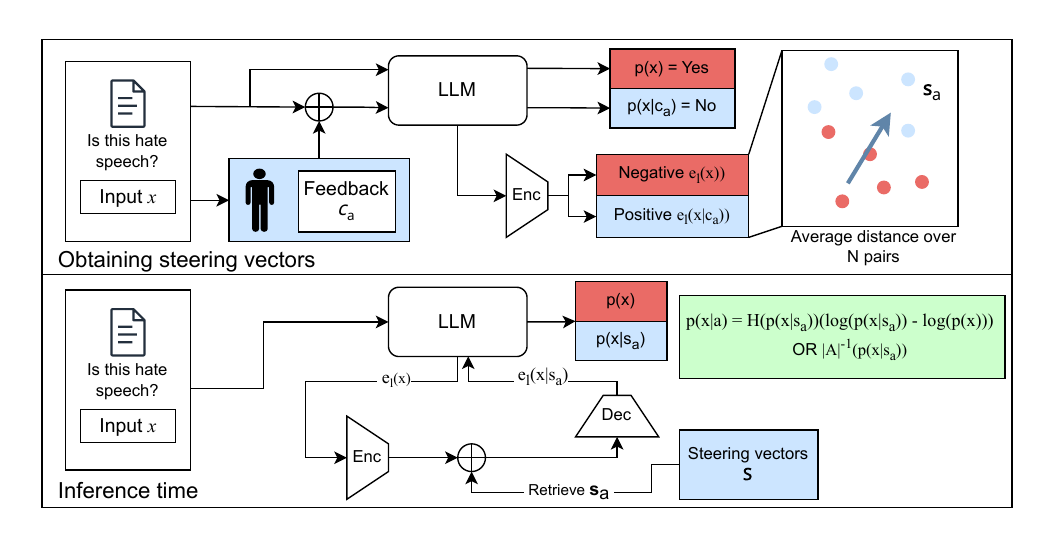}
    \caption{An illustration of our pipeline for one annotator $a$. For an input $x$ and feedback $c_a$, we obtain a contrastive pair as the LLM output with ($p(x|c_a)$) and without feedback ($p(x)$). Then, we take the SAE vector $s_a$ as the difference in the Sparse Auto-Encoder (SAE) representations, averaged across $N$ contrastive pairs. During inference, we add the SAE vector to the encoded representation of the input to enhance alignment at inference time.}
    \label{fig:pipeline}
    \vspace{-0.5em}
\end{figure*}

Our main contributions are as follows:
\vspace{-1mm}
\begin{itemize}

\vspace{-1mm}
    \item We propose to adapt language models to diverse preferences in a low-resource (i.e. sparse) setting using model steering with Sparse Auto-Encoders (SAEs). This pipeline adds flexibility for altering model behaviour without training, reducing the requirements for training data and allowing quicker adaptation to novel tasks.
    
\vspace{-1mm}
    \item We propose an adaptation of contrastive decoding, which we call pluralistic decoding (PD), to dynamically combine multiple perspectives at the decoding step. This strengthens predictions that diverge from the baseline distribution, enhancing the influence of minority preferences.
    
\vspace{-1mm}
    \item  We present experiments using human and synthetically-generated feedback, and empirically demonstrate our method improves alignment to in-context feedback over the base model. Additionally, we demonstrate SAE-based model steering can improve alignment with as few as 50 calibration samples.
\end{itemize}

\section{Alignment to Sparse Feedback}

\vspace{-1mm}
Our full pipeline is shown in \Cref{fig:pipeline}. We incorporate diverse expert feedback from multiple perspectives using a variation of contrastive decoding, which we call \textit{pluralistic decoding}. This can easily be applied to any black-box language model with exposed logits. However, changing the input query is often a limited model intervention. In preliminary experiments we found pluralistic decoding does not significantly change the model output. With access to the intermediate model residuals, we additionally explore model steering to predefined perspectives using Sparse Auto-Encoder (SAE) representations.

\vspace{-1mm}
\paragraph{Preliminaries.}

An LLM refers to a causal language model that samples some next token prediction $t\sim p(x)$ over the token space for an input $x = \{x_1, x_2, \dots, x_n\}$ of length $n$. As shown in \Cref{eq:transformer}, an LLM can be thought of as a composite function of $L$ transformer decoder blocks, where the representation of the input $f_l(x)$ at layer $l \in L$ is dependent on the residual of the previous layer. 
\vspace{-1mm}
\begin{equation}
\label{eq:transformer}
    p(x) = f_{L}\circ f_{L-1} \circ \dots f_1(x)))
\end{equation}

\vspace{-1mm}
\paragraph{Defining Feedback.} For an arbitrary perspective, represented as an annotator $a$, alignment is defined as conditioning the logit space on some natural language feedback $c_a$, i.e. $p(x|a) \equiv p(x|c_a)$. With LLMs, experts can improve performance with natural language feedback, which allows for more semantically rich feedback from domain experts. We distinguish between \textit{coarse} and \textit{granular} feedback. Coarse feedback refers to top-down guidelines for alignment (e.g.``detect harmful statements'') while granular feedback is specific to a particular sample (e.g.``this is targeting a population of people''). Granular feedback is easier to quantify, as it is more concrete than values or principles \cite{jiang2021can}. However, it is often more expensive to obtain this feedback, especially in high-stakes domains such as law, medicine, and finance \cite{luo2023towards}. 
We wish to explore \textit{data-sparse} methods where a domain expert can quickly specify their perspective to maximize their valuable insights.

\vspace{-1mm}

\paragraph{SAE model steering.}
SAEs are language model interventions that retrieve some intermediate representation $f_l(x)$ and encode it to a higher dimension representation $enc(f_l(x))$, optimizing a loss function $\mathcal{L}(dec(enc(f_l(x))), f_l(x)) = \mathcal{L}_\text{MSE} + \mathcal{L}_\text{Sparsity}$ that preserves information with a reconstruction loss $\mathcal{L}_\text{MSE}$ and encourages sparsity with L1 regularization, $\mathcal{L}_\text{Sparsity}$. By intervening in the intermediate layers and modifying the embedding space, the intervention cascades to the output distribution $p(x)$. If the SAE is able to capture a well-formed representation, then it would create some consistent alignment to an expert at inference. For a validation set of $N$ samples, we take the SAE vector $\mathbf{s}_a$ as shown in \Cref{eq:steering}. In practice, we encode each contrastive pair into the encoding space, taking the difference with and without feedback in the input, and averaging the differences over $N$ samples.

\vspace{-2mm}
\begin{equation}
\label{eq:steering}
    \mathbf{s}_a = \frac{1}{N}\sum^N_i \left(enc(f_l(x|c_a)) - enc(f_l(x))\right)
\end{equation}

\vspace{-2mm}

\paragraph{Pluralistic decoding.} Once we have multiple responses to the query from various perspectives, next is the non-trivial task of combining them into one answer. 
We adapt contrastive decoding \cite{li-etal-2023-contrastive, jin-etal-2024-dvd} to multiple generations, which we call \textbf{Pluralistic Decoding} (PD). 
\vspace{-2mm}
\begin{multline}
\label{eq:cd}
    p(x|A) = \softmax\sum_{a \in A} H(p(x|c_a))\\ \left((1+\alpha)\log(p(x|c_a))-\alpha\log(p(x))\right)
\end{multline}

\vspace{-1mm}
As defined in \Cref{eq:cd}, the final distribution $p(x|A)$ for a set of annotators $a \in A$ is defined as the sum of contrastive logits weighted by the entropy of their probabilistic distribution. Intuitively, this encourages a higher weighting for distributions with more uncertainty, or more plausible token predictions. When not using PD, we perform a simple mean of the logits, i.e. $p(x|A) = \frac{1}{|A|}\sum_{a \in A} p(x|c_a)$.





\section{Experimental Settings}

\vspace{-1mm}
\begin{table*}[t]

\renewcommand\arraystretch{1}

    
    \centering
    \setlength{\tabcolsep}{4pt}

    \resizebox{0.85\linewidth}{!}{

    \begin{tabular}{p{0.5cm} |p{4.5cm}|ccc|ccc|ccc}

    \toprule

        &\multirow{2}{*}{Setting}  & \multicolumn{3}{c|}{GQA}&\multicolumn{3}{c|}{LHS}& \multicolumn{3}{c}{MisLC}\\

        \cmidrule{3-11}
        &&Ma-f1$\uparrow$  & Mi-f1 $\uparrow$ & JS $\downarrow$ &Bin-f1$\uparrow$ & Ma-f1 $\uparrow$ & Mi-f1 $\uparrow$ &Bin-f1$\uparrow$  &Ma-f1 $\uparrow$ & Mi-f1 $\uparrow$ \\

      \midrule

    \multirow{6}{*}{\rotatebox[origin=c]{90}{\texttt{Llama3.1-8b}}}
        ~ & \texttt{Zero-shot} & 10.3 & 36.9 & 0.345 & 15.3 & 9.7 & 45.1 & 16.1 & 20.6 & 30.6 \\    
        ~ &\texttt{Few-shot, n=3} & 5.8 & 12.8 & 0.347 &\textbf{33.0} &\textbf{16.5} &\textbf{90.1} & 7.4 & 6.3 & 72.2 \\ 

        \cmidrule{2-11} 
        ~ &\texttt{Full feedback} + \texttt{PD} & \textbf{27.2} & \textbf{47.9} &\textbf{0.245} & 23.7 & 11.8 & 78.3 & \textbf{18.7} &\textbf{ 28.1 }& 32.9 \\ 
        \cmidrule{2-11}
        ~ &\texttt{SAE Vectors} ($N$=50) & 13.6 & 40.5 & 0.291 & 30.4 & 15.2 & 84.6 &16.9 & 11.0 & 71.8 \\ 
        ~ &\texttt{SAE Vectors} ($N$=50) + \texttt{PD} & 14.1 & 39.2 & 0.291 & 28.9 & 14.4 & 83.4 & 16.7 & 9.7 & 71.8 \\ 

             \midrule
             \midrule

    \multirow{6}{*}{\rotatebox[origin=c]{90}{\texttt{Gemma2-9b}}}
        ~ & \texttt{Zero-shot} & 8.9 & 27.9 & 0.414 & 36.8 & 18.4 & 81.1 & 12.6 & 6.3 & 73.4 \\   
        ~ &\texttt{Few-shot, n=3} & 0.9 & 1.2 & 0.290 & 46.3 & 23.2 & 88.7 & 5.1 & 2.6 & \textbf{75.6} \\ 
             
        \cmidrule{2-11} 
        ~ &\texttt{Full feedback} + \texttt{PD} & \textbf{26.4} & \textbf{46.2} & \textbf{0.239} & \textbf{77.7} & \textbf{38.8} & \textbf{94.0} & \textbf{18.9} & \textbf{ 9.4} & 71.1 \\         

        \cmidrule{2-11}
        ~ &\texttt{SAE Vectors} ($N$=50) & 11.3 & 42.4 & 0.289 & 21.4 & 11.4 & 49.9 & 17.8 & 8.9 & 67.5 \\ 
        ~ &\texttt{SAE Vectors} ($N$=50) + \texttt{PD} & 12.4 & 41.2 & 0.318 & 15.2 & 12.6 & 87.9 & 17.3 & 8.7 & 70.9 \\ 




      \bottomrule

    \end{tabular}

    }

    \caption{Summary of our experimental results across two models, \texttt{Llama3.1-8b} and \texttt{Gemma2-9b}. $\uparrow$ indicates higher is better, $\downarrow$ lower is better, and we highlight the best result per dataset and model in \textbf{bold}.}

    \label{tab:baseline}

    \vspace{-0.5em}

\end{table*}

\paragraph{Datasets}
Implementation details can be found in \Cref{app:experiments}. We use the following datasets:
\begin{itemize}

\vspace{-1mm}
    \item \textbf{GlobalOpinionQA (GQA)} \cite{durmus2023towards} --- We take the synthetic feedback from the mixed setting of \citet{feng-etal-2024-modular}. We report the Jensen-Shannon (JS) distance as per previous work, but we also report the macro- and micro-f1 scores on the ``majority'' opinion, taken as the $\argmax$ of each distribution. 
    
\vspace{-1mm}
    \item For domain-specific tasks, we test two datasets, \textbf{Legal Hate Speech (LHS)} and \textbf{Misinformation with Legal Consequences (MisLC)} \cite{luo2023towards, luo2024misinformation}. We take 3 annotator comments as granular feedback in MisLC, randomly sampled 5 axes of coarse feedback in LHS, and report f1 scores as per \cite{luo2023towards}. 
\end{itemize}

\vspace{-1mm}
\paragraph{Models and Baselines}
We experiment with \texttt{Llama3.1-8b} \cite{he2024llama} and \texttt{Gemma2-9b} \cite{team2024gemma}. We use the base model without instruction tuning as previous works \cite{feng-etal-2024-modular} found the unaligned version of the model adapted the best for pluralistic alignment. For SAEs, we select pre-trained SAEs from \cite{he2024llama, lieberum2024gemma}. For the tuning set $N$, we experiment with up to 150 samples but find that we have equal performance with $N=50$.

We implement the following baselines:
\begin{itemize}

\vspace{-1mm}
    \item \textbf{Zero-shot} --- prompting without feedback.
    
\vspace{-1mm}
    \item \textbf{Few-shot} --- a naive implementation of low-resource alignment. From the tuning set, we sample 3 random examples per input and insert them into the zero-shot prompt.
\end{itemize}
We implement three experimental settings: \textbf{Full feedback} setting with PD, and a sparse feedback setting where we consider \textbf{SAE Vector}-based model steering alone, as well as SAE steering with pluralistic decoding.
Please refer to \Cref{app:hyperparameters} for more details.

\section{Results and Discussion}

\vspace{-1mm}

\paragraph{Baseline results.} Our main results are shown in \Cref{tab:baseline}. The naive few-shot setting has poor performance overall, but performs surprisingly well on the LHS dataset. We believe that GQA and MisLC exhibit the expected behaviour — the few-shot setting is chosen to be naive and intended to have relatively poor performance as we only consider one annotator from the full set $A$. However, LHS's strong performance indicates the presence of lexical or morphological patterns that are easily learned over few-shot demonstrations, and the model likely bypasses the nuance of individual axes of feedback. This trend continues when adding more few-shot examples, as shown in \Cref{app:experiments}.

\begin{figure*}[t!]
    \centering
\begin{subfigure}{0.42\linewidth}
    \centering
    \includegraphics[trim=0cm 0.7cm 0cm 1cm, clip, width=0.85\linewidth, height=0.45\linewidth]{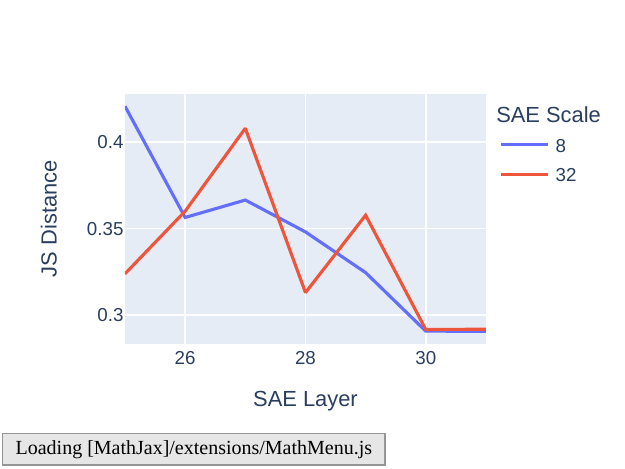}
    \caption{GQA dataset (lower is better).}
\end{subfigure}
\begin{subfigure}{0.42\linewidth}
    \centering
    \includegraphics[trim=0cm 0.7cm 0cm 1cm, clip, width=0.85\linewidth, height=0.45\linewidth]{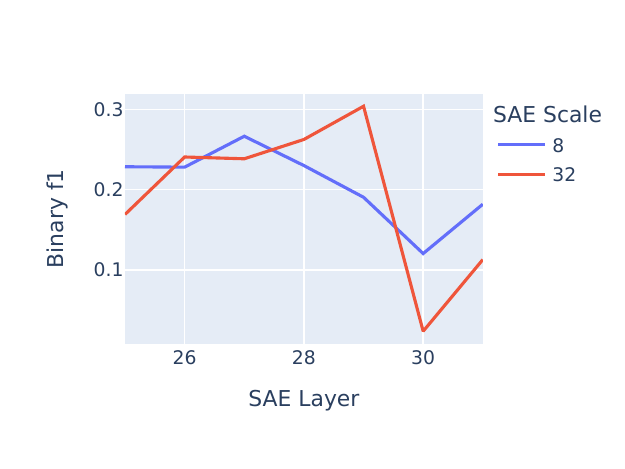}
    \caption{LHS dataset (higher is better).}

\end{subfigure}
\caption{The variance of the key performance metric for two datasets across different SAE scales and intervention layers for \texttt{Llama3.1-8b}. Please refer to \cref{app:experiments} for \texttt{Gemma2-9b} figures.}

\label{fig:sae-layer}
\vspace{-1mm}
\end{figure*}

\paragraph{Alignment scales with data sparsity.} For the full feedback setting, we use all available feedback in the gold labels, or synthetically generated with the same process as previous works \cite{feng-etal-2024-modular}, and combine the predictions into one distribution with pluralistic decoding (PD). This process gives a consistent performance boost over the zero-shot baseline, especially for the GQA dataset which decreases 0.10 points in JS distance for \texttt{Llama3.1-8b}. We believe this is because the GQA has the most dense feedback, since we are using synthetically generated feedback compared to manual annotations from lawyers.

On \texttt{Gemma2-9b}, the results are even stronger with full feedback. While the few-shot performance indicates the possibility of lexical artifacts in the LHS dataset, we observe PD improves performance beyond few-shot prompting for \texttt{Gemma2-9b}. With \texttt{Gemma2-9b} on MisLC, full feedback enhances performance on the positive class by 6.3 points f1 score. With the exception of \texttt{Gemma2-9b} performance on LHS, we observe relatively consistent improvements over the zero-shot baseline. This indicates there is some merit to implementing model steering for alignment, although full feedback is still preferable when possible.
\paragraph{Increased alignment for domain-specific tasks with sparse feedback.} In applications such as content moderation, experts are more concerned with understanding the potential utility of a system rather than its absolute performance \cite{masud-etal-2024-hate}, so precision is often a more important metric than f1 score. SAE model steering \textit{decreases the number of false positives} in the \texttt{Llama3.1-8b} experiments on both legal datasets without sacrificing accuracy on the positive class. However, the performance on the positive class is still low, indicating there is no change in the task understanding. 

We also do not observe improvements when combining model steering and pluralistic decoding in an end-to-end pipeline. We theorize this is because we are intervening on the transformer residual, we are straying too far from the learned space. For example, PPO places a KL divergence penalty in its training loss to prevent the model from straying too far from the pre-trained information. Since model steering is simple addition in the SAE vector space, we do not place such constraints. Previous works \cite{wang-etal-2025-beyond-prompt} demonstrate increasing the magnitude of the steering vector also results in nonsensical outputs, indicating there are limits to the degree of intervention.
\begin{figure}
    \centering
    \includegraphics[trim=0cm 0.9cm 0cm 1cm, clip, width=0.85\linewidth]{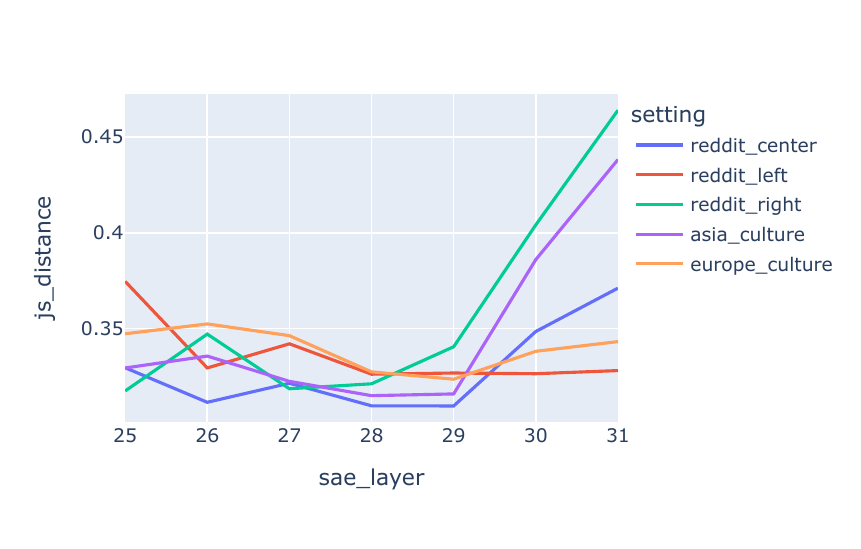}
    \caption{The per-layer JS distance of each steering vector for GQA using \texttt{Gemma2-9b}.}
    \label{fig:placeholder}
\end{figure}



\paragraph{Layer choice in SAE Vector steering.} We investigate the choice of SAE layer in \Cref{fig:sae-layer}. While there are some general trends in performance, there are also fluctuations and nonlinear trends that are likely affected by the underlying information stored at each layer of the transformer. These are also the layers that have the most fluctuations when we vary feedback. The best intervening layer also varies by task and model--- while GQA sees the best performance at layers 30 and 31 for \texttt{Llama3.1-8b}, we observe a drop in performance in the LHS and MisLC datasets. Also, the same is not true for \texttt{Gemma2-9b}. \Cref{fig:placeholder} shows that the best layer also varies by steering vector, and layers 30 and 31 actually have the worst performance for the steering vectors tuned to reddit\_right and asia\_culture.
\vspace{-1mm}

\begin{figure}[t!]
    \centering
    \includegraphics[trim=0cm 0.9cm 0cm 1cm, clip, width=0.8\linewidth]{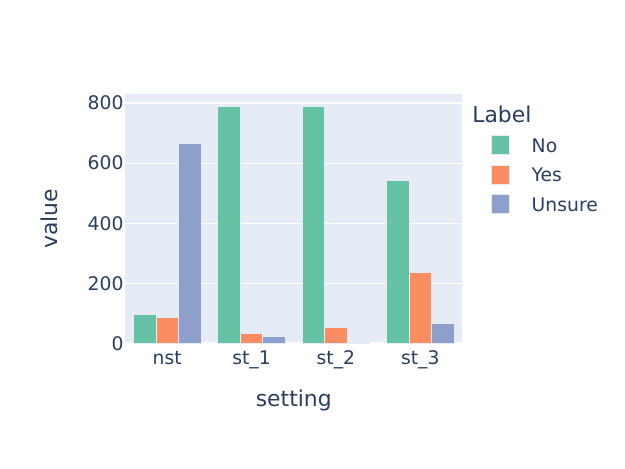}
    \caption{The label distribution of each class for the SAE vector setting on the MisLC dataset. nst refers to no steering, i.e. the zero-shot setting. st\_1, st\_2, and st\_3 refers to steering vector 1, 2, and 3 respectively.}
    \label{fig:lab_dist_mislc}
    \vspace{-1em}
\end{figure}

\begin{figure}[t!]
    \centering
    \includegraphics[trim=0cm 0.9cm 0cm 1cm, clip, width=0.8\linewidth]{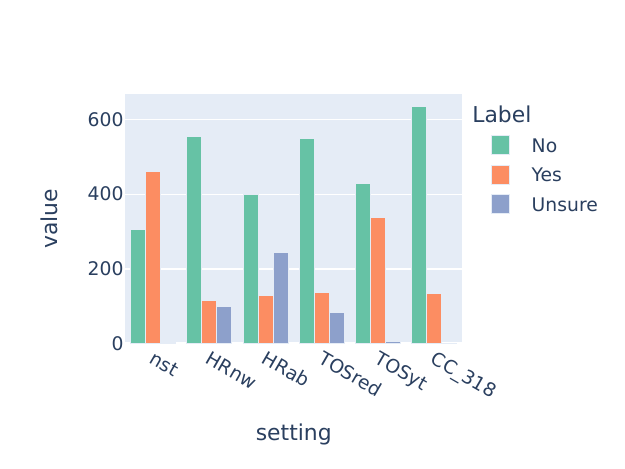}
    \caption{The label distribution of each class for steering on the LHS dataset for \texttt{Llama3.1-8b}. nst refers to no steering, i.e. the zero-shot setting.}
    \label{fig:lab_dist_lhs}
    \vspace{-1em}
\end{figure}
\paragraph{Alignment to diverse feedback.} 
We decompose the MisLC and LHS task to examine label predictions per steering vector at the chosen layer, as shown in \Cref{fig:lab_dist_mislc} and \Cref{fig:lab_dist_lhs}. The first two annotators give similar predictions but the third has more predictions to the positive class (Yes). This results in a higher recall but lower precision, which balances out to a similar Bin-f1 score. 
In the LHS task, there are three distinct groupings of hate speech definitions --- human rights laws (HR), social media policies (TOS), and criminal offences (CC). If our method is sufficiently aligned to the feedback, then the criminal code (CC) would predict the positive class the least frequently. We do find that one vector aligned to a social media policy (TOSyt) gives a higher positive rate, but the other vectors are at a similar level. 
One interesting observation is the negative class (No) is predicted the most on CC\_318, corresponding to the most severe definition of hate speech in the LHS dataset (Advocating Genocide). This demonstrates there is some alignment to the label distribution of the coarse feedback used to generate the steering vector. We believe the steering vectors improve alignment to the stricter legal definitions compared to the base model. Please refer to \Cref{app:experiments} for comparison of coarse-grained and fine-grained feedback on one dataset.


\section{Related Work}
\vspace{-1mm}
\paragraph{Alignment} Alignment is the general field of study towards ensuring AI aligns with human values \cite{ngo2022alignment}. It is an active area of research with many open questions: first, it is somewhat difficult to know how to define all possible dimensions of human values \cite{sorensen2024roadmap}. 
For example, an LLM agent learns to lie or deceive the user because it was never rewarded for honesty \cite{ngo2022alignment, williams2024targeted}. 


\vspace{-1mm}
\paragraph{Steering} Model steering is an emerging field of study on directing language model behaviour by its own internal representations, rather than updating parameters through training \cite{zou2023representation}. Sparse Autoencoders (SAEs) are a promising direction for model steering. Previous works demonstrate SAEs produce more interpretable features that detangle polysemantic representations \cite{cunningham2023sparse, chalnev2024improving}, and improve the ability to steer granular behaviours \cite{zhao-etal-2025-steering}. There are works that have successfully applied steering to control knowledge selection \cite{zhao-etal-2025-steering} or helpfulness \cite{chalnev2024improving}, but none have tried steering over multiple dimensions.
Please refer to \Cref{app:related} for more related work. 
\section{Conclusion} 
In this work, we explore how to increase AI alignment to a diverse range of human values. We introduce pluralistic decoding and utilize model steering via sparse auto-encoders to increase the signal of human feedback in low-resource settings. We empirically demonstrate that model steering offers consistent improvement over the baseline for 50 annotated samples, and demonstrate a decrease in false positives on domain-specific tasks. 
However, this method does not enhance the task understanding or underlying model reasoning. 
We hope this work highlights the importance of pluralistic alignment and inspires future work in the area. 
\newpage
\section*{Limitations}
There are several limitations to our work. First, we do not evaluate the performance of our system in the presence of noise, so this does not study how the consistency of the annotator bias affects the clarity of the steering vector. Also, while we propose our methodology with SAEs, this could theoretically be applied directly to the transformer residual, but we do not test the efficacy.

Sparse Auto-Encoders are a very recent development for interpretability and they are still highly experimental. There are concerns about scalability to larger models, for example. Our method assumes there are already SAEs trained for a specific language model to a certain level of reconstruction accuracy --- if not already available, an SAE would need to be trained from scratch, which is arguably more expensive than parameter-efficient tuning.

As briefly mentioned, SAE steering vectors produce more nonsensical outputs compared to plain pluralistic decoding. Because we propose to intervene in the intermediate layers of a transformer, there is a risk of straying too far from the language models' learned activation space. This could lead to language models becoming more susceptible to jailbreaks among other vulnerabilities. We urge further research in this direction. They also do not interact well with other modifications to the base model, such as alignment tuning or the pluralistic decoding tested in this paper.

\section*{Ethics Statement}
This paper explores alternatives to prompting and training for aligning model behaviour in a low-resource setting. We share the findings for the NLP community, and we will release the code for the purpose of further scientific exploration. In terms of utility, we believe this could one day lead to more customizable language models; for example, you could store these steering vectors in the same infrastructure as vector databases currently used for Retrieval-Augmented Generation, and retrieve a specific vector for certain tasks in place of, eg. a system prompt or LoRA adapter.
\section*{Acknowledgements}
The research is in part supported by the NSERC Discovery Grants and the Research Opportunity Seed Fund (ROSF) of Ingenuity Labs Research Institute at Queen's University.
\bibliography{anthology,custom}

\newpage
\appendix
\section{Extended Related Work}
\label{app:related}
\paragraph{Alignment} Alignment is the general field of study towards ensuring AI aligns with human values \cite{ngo2022alignment}. It is an active area of research with many open questions: first, it is somewhat difficult to know how to define all possible dimensions of human values \cite{sorensen2024roadmap}. 
While it seems to be a closed environment, there are many dimensions of human value that are irrevocably intertwined with society, so it is difficult to articulate them for artificial intelligence \cite{jiang2021can}. 
This is an issue when, for example, an LLM agent learns to lie or deceive the user in order to maximize its goals because it was never rewarded for honesty \cite{ngo2022alignment, williams2024targeted}. 
The most popular methods for model alignment include training and in-context prompting \cite{ovadia2023fine}, but these methods are still imperfect guides.
For example, prompting language models to align to certain personas (eg. a doctor, teacher) can have extremely unpredictable effects on performance \cite{zheng-etal-2024-helpful}. 

\paragraph{Contrastive Decoding} 
Prompting in the context of LLMs implies hard prompting, or appending extra context as additional hard tokens to the input \cite{liu2023pre}. 
A popular variation is retrieval-augmented generation (RAG), where there is an external retriever that finds the most relevant document to enhance the LLM's knowledge \cite{gao2023retrieval}. 
Contrastive decoding has emerged as a method to further enhance the knowledge in a language model's logits --- by taking an ``expert'' and ``amateur'' probabilistic distribution to the same answer \cite{li-etal-2023-contrastive}.
\citet{jin-etal-2024-dvd} enhances RAG over multiple documents by contrasting the highest- and lowest-scoring document. However, these are only two logits from the entire set, and it is poorly suited for pluralistic alignment.

\paragraph{Steering} Model steering is an emerging field of study on directing language model behaviour by its own internal representations, rather than updating parameters through training \cite{zou2023representation}. They have been shown to produce more interpretable features that detangle polysemantic representations in the model weights \cite{cunningham2023sparse, chalnev2024improving}, and also improve the ability to steer granular behaviours \cite{zhao-etal-2025-steering}. There are works that have successfully applied steering to control knowledge selection \cite{zhao-etal-2025-steering} or helpfulness \cite{chalnev2024improving}, but none have tried steering to multiple dimensions.

\section{Extended Experimental Details}
\label{app:hyperparameters}
\subsection{Datasets}
All datasets are licensed for public research use, which is consistent with the purpose of this work. While the datasets were pre-anonymized, we also manually inspect a few samples and remove any personal information such as usernames or annotator names.

We analyze our method on the following datasets:
\begin{itemize}
    \item GlobalOpinionQA (GQA) \cite{durmus2023towards} --- to assess distributional alignment. We report the Jenson-Shannon (JS) distance as per previous work \cite{feng-etal-2024-modular}, but we also report the macro- and micro-f1 scores on the ``majority'' opinion, taken as the $argmax$ of each distribution. They fine-tune \texttt{Mistral-7b} with LoRA find the best performance by sampling six unique perspectives for each input prompt.
    \item Legal Hate Speech (LHS) and Misinformation with Legal Consequences (MisLC) \cite{luo2023towards} --- To analyze the performance on more specialized tasks, we test two datasets from the legal domain for hate speech and misinformation detection. This series of datasets has fine-grained annotations based on several legal definitions at varying levels of severity.
\end{itemize}

\paragraph{GlobalOpinionQA} We take the small split of GlobalOpinionQA, with 5,752 samples from all countries, and randomly sample 200 as tuning for the steering vector. We also use their method to generate synthetic feedback for every sample, and perform pluralistic decoding to enhance the dense feedback from every perspective. We also use their evaluation code, where they isolate specific tokens that map to each multiple choice answer from the top 10 and perform softmax on the gathered probabilities.

\paragraph{LHS and MisLC} We use the train-test split provided by the authors, which contains ~1300 training samples (gold data and some sampled silver data), and 709 test samples. We sort the training set by the amount of written feedback and take the top 50. There were approximately 39 samples with written feedback, leaving the other 11 as randomly sampled. Following previous work, we take the top predicted token to simulate greedy sampling and count invalid tokens as a separate prediction (-1), which we choose to map to the Unsure class (2). 

We report metrics following \cite{luo2024misinformation}, with binary f1 on the positive class and macro-f1 statistics considering classes 1 and 2 as positive.
For feedback, we use the notes and comments from legal annotators available in the respective datasets. For the misinformation dataset, this includes sources to verify the veracity of statements, and for hate speech it contains more commentary/analysis on the sample itself.

\subsection{Prompts}
\label{app:prompts}
Our prompt templates are shown in \Cref{tab:prompts}. Please note that there is no space and we sample immediately after the colon --- adding a space can dramatically change the results. We make best efforts to match the prompt from the original work the dataset was sourced from.

For \texttt{Gemma2-9b}, we observed in preliminary experiments that the model tended to output an HTML <strong> tag instead of immediately predicting the answer, so we additionally append this tag to the prompt template. 
\begin{table}[ht]
    \centering
    \resizebox{\linewidth}{!}{
    \begin{tabular}{p{1.3cm}|p{6cm}}
        \toprule
        \textbf{ID}   & \textbf{Instruction} \\
        \midrule
\texttt{GQA}    & You are from the country of \{country\}, respond to the following instruction (with the help of a passage. Passage: \{feedback\_fine\}) \\& \\&
\{input\} \\& \\&
Answer:\{output\}
\\ \hline

\texttt{MisLC} &  From a legal perspective, misinformation can be problematic due to: \{feedback\_coarse\}\\&Claim: \{input\}\\&Does this claim contain misinformation? Answer Yes, No, or Unsure.\\&\\&Thinking: \{feedback\_fine\}\\&Answer:\{output\}                                                                                       \\ \hline
\texttt{LHS}& Hate speech policy: \{feedback\_coarse\}\\&Post: \{input\}\\&Does this post violate the above hate speech policy? Answer Yes, No, or Unsure.\\&\\&Thinking: \{feedback\_fine\}\\&Answer:\{output\}                                                                              \\
        \bottomrule
    \end{tabular}}
    \caption{Prompt templates for each dataset.\{\} indicates a part of the prompt that was substituted with the data, and we sample the token at \{output\}. Coarse feedback refers to task definitions, granular feedback refers to per-sample annotations from either human experts or fine-tuned models.}
    \label{tab:prompts}
\end{table}

\subsection{Implementation details}
\label{app:hyperparams}
All the experiments were run on either two Nvidia Quadro RTX 8000 GPUs with $46$ GiB of VRAM per GPU, or two Nvidia A100-PCIEs with $41$ GiB of VRAM per GPU. We use HuggingFace datasets for preprocessing; PyTorch, TransformerLens for hooking into the transformer activations; and SAELens for loading the SAEs. 

Since we operate directly on the probabilities rather than sampling, all experiments are deterministic, and we only report one run. Where applicable, we take the temperature to be 0.4 to simulate sampling. For $\alpha$, we chose a value of 0.2 following \cite{jin-etal-2024-dvd}.

\section{Additional Experiments and Analysis}
\label{app:experiments}
\begin{figure*}[t!]
    \centering
\begin{subfigure}{0.3\linewidth}
    \centering
    \includegraphics[trim=0cm 0.7cm 0cm 1cm, clip, width=\linewidth]{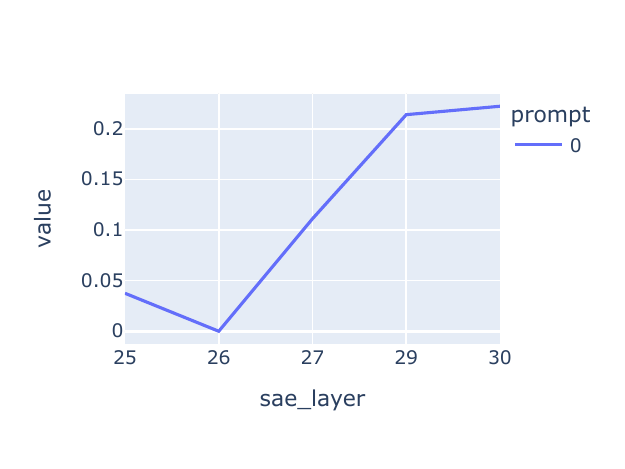}
    \caption{LHS dataset (higher is better).}
\end{subfigure}
\begin{subfigure}{0.3\linewidth}
    \centering
    \includegraphics[trim=0cm 0.7cm 0cm 1cm, clip, width=\linewidth]{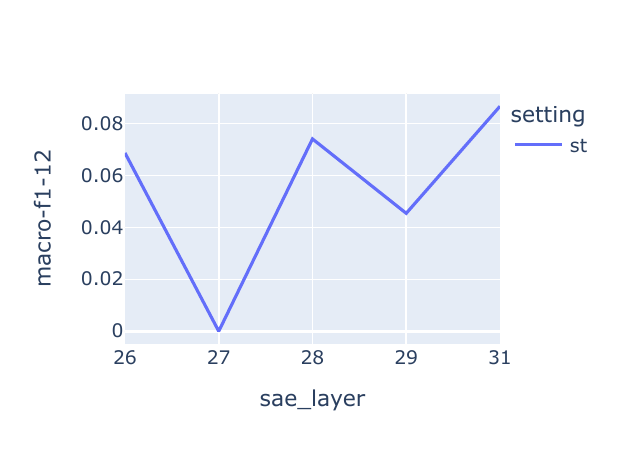}
    \caption{MisLC dataset (higher is better).}

\end{subfigure}
\begin{subfigure}{0.3\linewidth}
    \centering
    \includegraphics[trim=0cm 0.7cm 0cm 1cm, clip, width=\linewidth]{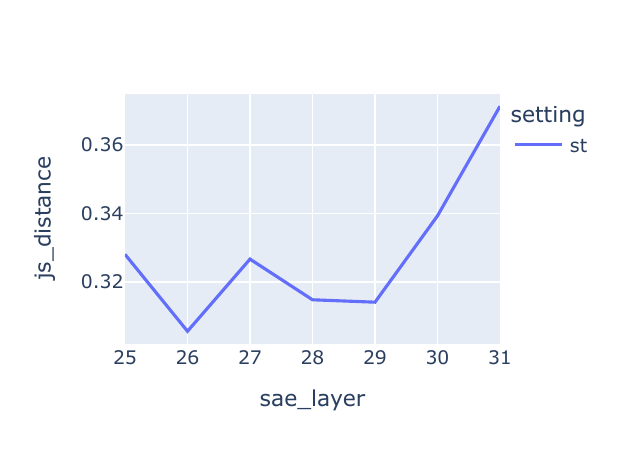}
    \caption{GQA dataset (lower is better).}

\end{subfigure}
\vspace{-1mm}
\caption{The variance of the key performance metric for three datasets across different intervention layers for \texttt{Gemma2-9b}.}

\label{fig:sae-layer-gemma}
\end{figure*}

\begin{figure}[t!]
\begin{subfigure}[c]{\linewidth}
    \centering
    \includegraphics[trim=0cm 0.7cm 0cm 1cm, clip, width=0.8\linewidth, height=0.4\linewidth]{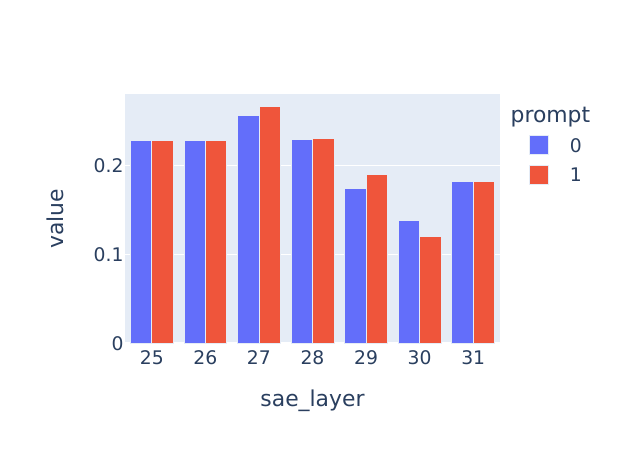}
    \caption{8x SAE.}
\end{subfigure}
\begin{subfigure}[c]{\linewidth}
    \centering
    \includegraphics[trim=0cm 0.7cm 0cm 1cm, clip, width=0.8\linewidth, height=0.4\linewidth]{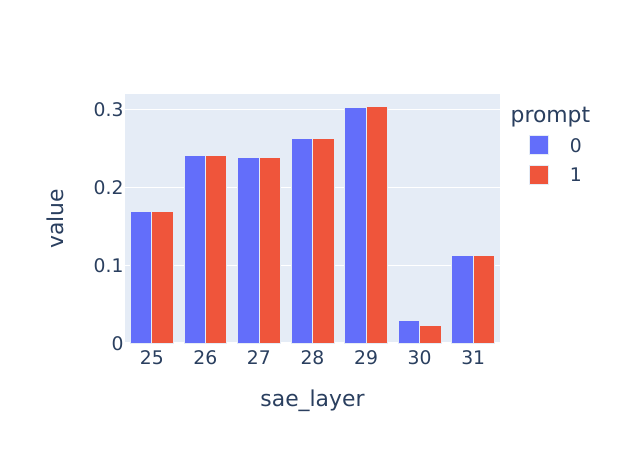}
    \caption{32x SAE.}

\end{subfigure}
\caption{The variation of f1 score on the LHS dataset when steered on coarse (0) vs. granular (1) feedback.}
\label{fig:feedback_var}
\end{figure}

This section details additional figures to supplement the core results. 
\begin{itemize}
    \item \textbf{Coarse vs. Granular feedback.} \Cref{fig:feedback_var} shows the variation in performance when steering to diverse coarse feedback vs. granular feedback. As shown, the variation is relatively minimal --- while the performance fluctuates between layers, especially in the 8x expansion SAE, there is no clear pattern to which is better and which is worse.
    \item \textbf{Extended SAE Vector results.}
We additionally present more results that were omitted from the main body due to space constraints. This includes per-layer SAE Vector results for \texttt{Gemma2-9b} in \Cref{fig:sae-layer-gemma}. Additionally, we further compare the two sizes of SAE vector available for \texttt{Llama3.1-8b}

\end{itemize}

\begin{table}[!ht]
    \centering    
    \setlength{\tabcolsep}{2pt}

    \resizebox{0.8\linewidth}{!}{
    \begin{tabular}{l|l|l|l|l|l}
\toprule
 setting & p-1 & r-1 & f1-1 & macro-f1-12 & micro-f1-12 \\ \midrule
 nst & 8.7 & 72.7 & 15.5 & 7.7 & 43.9 \\ 
 HRnw & 22.2 & 47.3 & 30.2 & 15.1 & 72.9  \\ 
 HRab & 21.7 & 50.9 & 30.4 & 15.6 & 53.6  \\ 
 TOSred & 21.2 & 52.7 & 30.2 & 15.1 & 72.5 \\ 
 TOSyt & 12.4 & 76.4 & 21.4 & 10.7 & 59.8 \\ 
 CC\_318 & 22.1 & 54.5 & 31.4 & 15.7 & 83.0  \\ 
 \bottomrule
    \end{tabular}
    }
    \caption{Per-vector performance on the LHS dataset for \texttt{Llama-3.1-8b}.}
    \label{tab:hs}
\end{table}

\begin{table*}[t]

\renewcommand\arraystretch{0.8}

    \centering

    \setlength{\tabcolsep}{4pt}

    \resizebox{0.6\linewidth}{!}{

    \begin{tabular}{p{4cm}|ccc|ccc}

    \toprule


\multirow{2}{*}{Model}&\multicolumn{3}{c|}{\texttt{Llama3.1-8b}}&\multicolumn{3}{c}{\texttt{Gemma2-9b}} \\
\cmidrule{2-7}
          &Bin-f1$\uparrow$ &Ma-f1$\uparrow$  & Mi-f1 $\uparrow$ &Bin-f1$\uparrow$ &Ma-f1$\uparrow$  & Mi-f1 $\uparrow$\\
        \midrule
        \texttt{Zero-shot} &16.1&20.6&30.6&12.6&6.3&73.4\\
        \texttt{SAE Vector} ($N$=50) &16.9&11.0&71.8&17.8&8.9&67.5\\
        \texttt{Annotator 1} & 14.7 & 10.5 & 73.9 & 11.0 & 5.5 & 72.9 \\ 
        \texttt{Annotator 2} & \textbf{18.5} & 10.7 & \textbf{74.1} & 16.2 & 8.1 & 50.9 \\ 
        \texttt{Annotator 3} & 17.2 & \textbf{13.6} & 59.8 & 13.6 & 6.8 & 73.3 \\ 

         \bottomrule

    \end{tabular}

    }
        \caption{Further investigation into the MisLC dataset performance per fine-grained steering vector.}  

    \label{tab:mislc_perf}

\end{table*}



\begin{figure}[t!]
\begin{subfigure}[c]{\linewidth}
    \centering
    \includegraphics[trim=0cm 0.7cm 0cm 1cm, clip, width=0.88\linewidth]{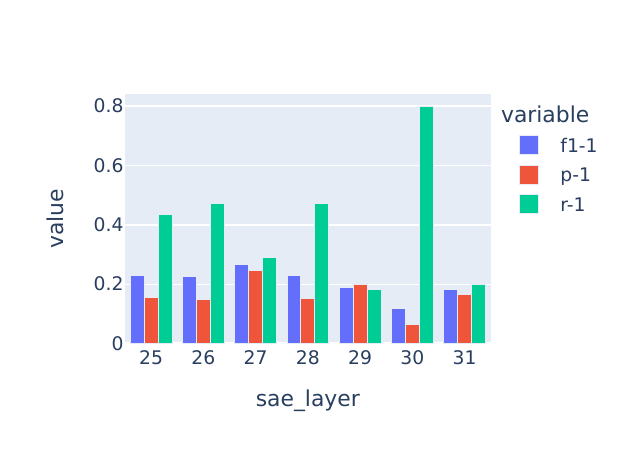}
    \caption{8x SAE.}
\end{subfigure}
\begin{subfigure}[c]{\linewidth}
    \centering
    \includegraphics[trim=0cm 0.7cm 0cm 1cm, clip, width=0.88\linewidth]{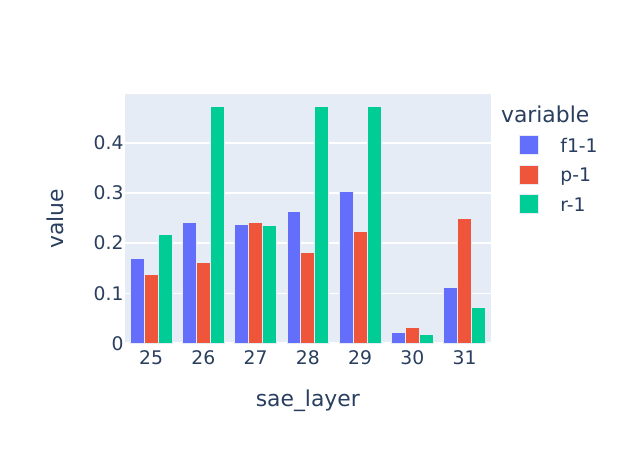}
    \caption{32x SAE.}

\end{subfigure}
\caption{The variation of f1, precision, and recall on the LHS dataset at different SAE layers and scales. }
\label{fig:prf1}
\end{figure}

\begin{table}[t]

\renewcommand\arraystretch{1}

    
    \centering
    \setlength{\tabcolsep}{4pt}

    \resizebox{\linewidth}{!}{

    \begin{tabular}{p{2.5cm}|ccc|ccc}

    \toprule

        \multirow{2}{*}{Setting}  &\multicolumn{3}{c|}{LHS}& \multicolumn{3}{c}{MisLC}\\

        \cmidrule{2-7}
        &Bin-f1$\uparrow$ & Ma-f1 $\uparrow$ & Mi-f1 $\uparrow$ &Bin-f1$\uparrow$  &Ma-f1 $\uparrow$ & Mi-f1 $\uparrow$ \\

      \midrule


    \texttt{n=5} & 52.5 & 26.3 & 92.9 & 14.9 & 8.2 & \textbf{75.8} \\ 
    \texttt{n=10}  &\textbf{53.8} &\textbf{26.9} &\textbf{93.1} & 11.5 & 5.8 & 75.1 \\ 


      \bottomrule

    \end{tabular}

    }

    \caption{Additional few-shot results on LHS and MisLC for \texttt{Llama3.1-8b}.}

    \label{tab:fewshot}

    \vspace{-0.5em}

\end{table}

\end{document}